%% file: main.tex
\documentclass[letterpaper, 10 pt, conference]{ieeeconf}
\IEEEoverridecommandlockouts
\input{packages}

\newcommand{\methodname}{\textsc{CAVER}}

\title{\methodname{}: Curious Audiovisual Exploring Robot}

\author{Luca Macesanu$^{*}$, Boueny Folefack$^{*}$, Samik Singh$^\dagger$, Ruchira Ray$^\dagger$, \\
Ben Abbatematteo, Roberto Mart\'in-Mart\'in
\thanks{$^*, \dagger$ denotes equal contribution. All authors are with The University of Texas at Austin}}

\begin{document}
\maketitle
\thispagestyle{empty}
\pagestyle{empty}

\input{00_abstract}
\input{01_intro}
\input{02_related}

\input{03_method}
\input{04_experiments}
\input{05_conclusion}
\footnotesize
\bibliographystyle{IEEEtran}

\bibliography{IEEEabrv, refs}
\end{document}

%% file: packages.tex
\usepackage{times}
\usepackage{cite}
\usepackage{amsmath,amssymb,amsfonts}
\usepackage[font=small]{caption}
\usepackage{subcaption}
\usepackage{algorithmic}
\usepackage{graphicx}
\usepackage{textcomp}
\usepackage{xcolor}
\usepackage{booktabs}

\usepackage{lipsum}
\usepackage{hyperref}
\usepackage[export]{adjustbox}

%% file: 00_abstract.tex
\begin{abstract}
Multimodal audiovisual perception can enable new avenues for robotic manipulation, from better material classification to the imitation of demonstrations for which only audio signals are available (e.g., playing a tune by ear).
However, to unlock such multimodal potential, robots need to learn the correlations between an object's visual appearance and the sound it generates when they interact with it.
Such an active sensorimotor experience requires new interaction capabilities, representations, and exploration methods to guide the robot in efficiently building increasingly rich audiovisual knowledge.
In this work, we present \methodname{}, a novel robot that builds and utilizes rich audiovisual representations of objects.
\methodname{} includes three novel contributions: 1) a novel 3D-printed end-effector, attachable to parallel grippers, that excites objects' audio responses, 2) an audiovisual representation that combines local and global appearance information with sound features, and 3) an exploration algorithm that uses and builds the audiovisual representation in a curiosity-driven manner that prioritizes interacting with high uncertainty objects to obtain good coverage of surprising audio with fewer interactions.
We demonstrate that \methodname{} builds rich representations in different scenarios more efficiently than several exploration baselines, and that the learned audiovisual representation leads to significant improvements in material classification and the imitation of audio-only human demonstrations. More information: \url{https://robin-lab.cs.utexas.edu/CAVER}
\end{abstract}

%% file: 01_intro.tex
\section{Introduction}


Humans learn and exploit multimodal audiovisual cues in everyday life to obtain a more complete understanding of their environment and broader manipulation capabilities. 
We routinely fuse audio and vision to understand materials and reproduce behaviors: tapping a mug reveals glass vs. ceramic, and hearing a melody lets a musician find the right key. 
Building similar capabilities in robots would increase their robustness and autonomy, but requires a representation that couples \emph{how things look} with \emph{how they sound when interacted with}, and a way to acquire that representation efficiently through interaction.

\begin{figure}[t!]
    \centering
\includegraphics[width=1\columnwidth]{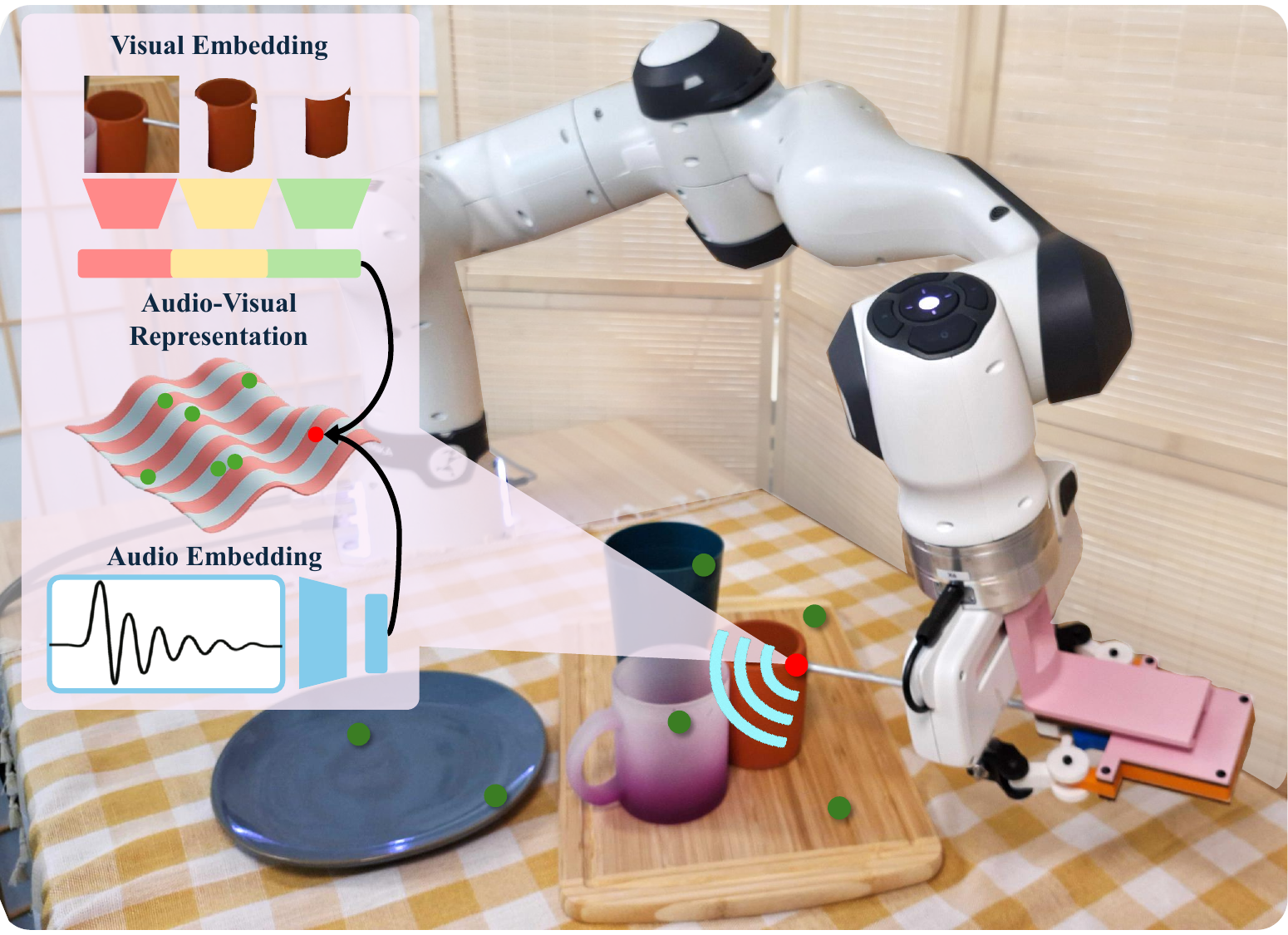}
    \caption{\textbf{Curiously Building and Exploiting an Audiovisual Representation with \methodname{}}. \methodname{} incrementally builds a KNN-based audiovisual representation capable of audio-to-visual and visual-to-audio prediction. To explore an environment, \methodname{} considers many candidate interaction points (dots), and ranks their uncertainty by comparing their visual features to those of prior samples. \methodname{} then collects an audio sample of the most uncertain point (red) using our novel impact tool. The audio and visual features are added as a pair to the audiovisual representation in an intrinsically motivated process that results in efficient interactive exploration.}
    


    \label{fig:pull}
    \vspace{-2em}
\end{figure}


Underlying the multimodal audiovisual abilities of humans is a rich developmental process in which children interact with and explore the physical world around them~\cite{piaget1952origins}. 
When children play with objects, they generate multimodal stimuli from which they learn audiovisual correlations in a self-supervised manner~\cite{smith2005development}. 
Their exploration is incremental and, critically, \textit{curious}: children prioritize unfamiliar experiences so as to quickly and efficiently learn the audiovisual properties of the world~\cite{loewenstein1994psychology, jirout2012children}. 
While prior work in machine learning and robotics has explored how to build such audiovisual representations~\cite{owens2016visually, chen2024action2sound,clarke2022diffimpact, gao2023objectfolder}, they rely on previously acquired large datasets of visual and audio pairs collected manually, largely failing to account for how a robotic agent might explore an environment autonomously and efficiently.
In this work, we introduce \methodname{}, \textbf{C}urious \textbf{A}udio\textbf{v}isual \textbf{E}xploring \textbf{R}obot, a robot that autonomously builds and exploits an audiovisual representation of objects (Fig.~\ref{fig:pull}).
\methodname{} integrates a novel, custom-built impact tool that enables a robot to ``tap'' on object surfaces to elicit an audio response.
\methodname{} draws inspiration from human curiosity to guide the robot through interaction with  uncertain objects, first actively exploring visually unfamiliar objects to learn correlations between their appearance and impact acoustics, then using those correlations for downstream tasks.
Concretely, a strategy of uncertainty-guided sampling chooses where to strike next to generate a visual~\cite{ren2024groundedsamassemblingopenworld,resnet} and auditory~\cite{kubichek1993mel} feature representation of the interaction point, while a simple but effective k-nearest-neighbor (KNN) methodology on the audiovisual features enables immediate retrieval in either direction  (vision$\rightarrow$audio and audio$\rightarrow$vision) to exploit the information for manipulation (see Sec.~\ref{sec:exp4}).


We test \methodname{} by exploring various household scenes and incrementally building its audiovisual representation, and evaluate it on downstream tasks including audio prediction, material classification, and audio-based imitation. 
Across several household environments, \methodname{} enables rapid learning of object visual and acoustic properties in an autonomous and self-supervised manner. 
Through our experiments, we demonstrate that \methodname{}'s curious, uncertainty-driven exploration strategy yields a marked improvement in audio prediction quality over several sampling and exploration alternatives, and its audiovisual representation achieves 87{\%} accuracy in material classification, 66{\%} accuracy in audio-based imitation of melodies, and 42\% accuracy in action recognition (compared to a 27\% accurate human baseline).


\begin{figure*}[t!]
    \centering
\includegraphics[width=\textwidth]{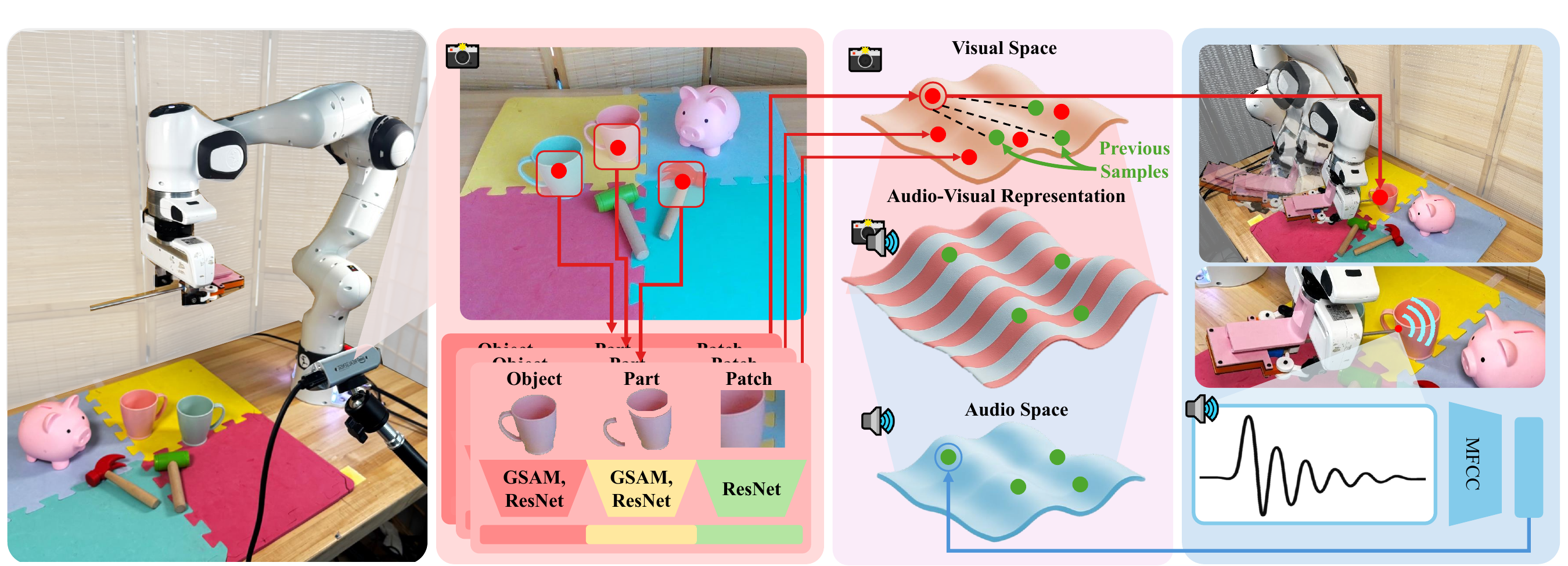}
    \caption{\textbf{Overview of \methodname{}'s audiovisual representation and curious exploration.} \methodname{}  curiously and efficiently learns correlations between object visual appearance and acoustic properties. Given candidate hitting points, \methodname{} uses a KNN model with fine tuned features from foundation vision models to predict corresponding impact sounds. To select informative interaction points, \methodname{} selects the most uncertain candidate hitting location using distance in visual feature space between a candidate (red) and all prior samples (green) as a proxy for uncertainty. After sampling the most uncertain candidate, the corresponding visual and audio embeddings are paired as two sides of the Audio-Visual representation. This can best be thought of as a bi-directional mapping where audio features can be used to predict visual features, visual features can be used to predict audio features, and concatenating the embeddings gives an informative multimodal representation of that sample point.}
    \label{fig:method}
    \vspace{-1em}
\end{figure*}

In summary, in \methodname{} our contributions include: 1) a reproducible, 3D-printed \emph{impact end-effector}; 2) a \emph{multi-scale audiovisual representation} combining visual and audio features in a KNN model, and learned feature weights aligning visual to audio distance; 3) an \emph{uncertainty-guided exploration} policy that prioritizes visually distinct object parts using farthest-first selection; and 4) a demonstration that this representation is immediately \emph{useful} across tasks---audio prediction, material classification, musical imitation, and audio-only manipulation inference---without large-scale pretraining on paired data.

%% file: 02_related.tex
\section{Related Work}
\methodname{} is a robot that can perceive, curiously explore, and quickly learn an audiovisual representation that can predict sound based on images and be applied to solve audio-related downstream tasks.
In this section, we contrast it to representative prior work that explored multimodal robot learning, interactive perception, and action sound generation. 

\textbf{Multimodal and Audiovisual Robot Learning.}
Leveraging multi-modal data in robotic tasks is less explored than the conventional vision-based learning setting~\cite{levine2016end, tang2024deep, o2024open, khazatsky2024droid}.
While several methods have combined vision with tactile information~\cite{chebotar2014learning,lee2019making} or force feedback~\cite{martin2017cross,liu2024forcemimic} for increased performance, leveraging audiovisual signals remains relatively underexplored.
Only a few methods have explored leveraging audio in imitation learning~\cite{ liu2024maniwav}, to handle occlusions~\cite{du2022play}, or as supervision for skill learning~\cite{thankaraj2023sounds}.
This primarily stems from a lack of large annotated audio datasets, as well as the inherent complexity of working with audio, including source ambiguity, environmental variability, and temporal dynamics.
The ObjectFolder benchmark~\cite{gao2022objectfolder, gao2023objectfolder} provides both synthetic and real impact sounds collected manually from a wide variety of household objects. 
DiffImpact~\cite{clarke2022diffimpact} explores differentiable rendering approaches to generating impact sounds, decomposing sounds into force profiles, and object acoustic properties.  
Still, none of these methods answer the question of how we can enable a robot to autonomously gather data to learn about the acoustic properties of the world around it. \methodname{} is able to autonomously generate its own dataset of audiovisual samples with which to learn a representation and solve downstream tasks.

\textbf{Interactive Perception.}
\methodname{} is a robot that learns to perceive audiovisual information by leveraging interactions with the environment in the form of hitting actions that reveal audio signals.
This is a form of \textit{interactive perception}~\cite{bohg2017interactive}, a family of techniques that enrich robots' perceptual capabilities by leveraging physical interactions. Examples include interactively segmenting a scene~\cite{mishra2009active,kenney2009interactive,kontogianni2023interactive}, perceiving kinematics~\cite{martin2014online,abbatematteo2019learning}, and recognizing  objects~\cite{maiettini2017interactive,martinez2017object}. 
Some interactive perception works~\cite{sinapov2011interactive, torres2005tapping, nguyen2018autonomous} demonstrate material property inference by interactively generating audio data. 
However, none of these methods provides an account of curiosity-driven exploration of the audiovisual properties of a scene or a sufficiently general representation that can be applied to diverse downstream tasks. 

\textbf{Curious Exploration.}
Curiosity as an intrinsic metric to guide exploration has proven its utility in reinforcement learning domains \cite{pinto2016curiousrobotlearningvisual,pmlr-v139-sontakke21a,pathak2017curiositydrivenexplorationselfsupervisedprediction}. Prior works have been able to build accurate dynamics representations of their environment by exploring state action pairs whose effects are unknown \cite{pmlr-v139-sontakke21a}. \methodname{} extends this approach to the audiovisual domain by exploring uncertain regions of the environment to learn their audio properties.

\textbf{Action Sound Generation.}
Previous approaches have attempted to predict the audio that an interaction would create based on the visual appearance of the interacted area.
Several methods use physics simulation to create virtual audio of interactions~\cite{jin2025avdar,liu2025haae,su2023physicsdrivendiffusionmodelsimpact}.
These methods require detailed models of the objects (e.g., shape, material), which are costly to obtain in unstructured environments.
Owens et al.~\cite{owens2016visually} introduced the idea of visually indicated sounds, audio signals that can be predicted based on the visual appearance of the interacted objects. 
Since then, several works have explored generating audio data conditioned on video, typically using large, manually curated datasets of humans interacting with objects~\cite{owens2018audio, chen2020generating, iashin2021taming, luo2023diff, su2023physics, du2023conditional, chen2024action2sound}. 
Despite the ability of these methods to generate plausible sounds accompanying videos, they fail to address the active exploration problem required to guide an agent to autonomously acquire its own audiovisual representation of the environment in the real world.
Moreover, many of these methods typically predict audio based on observed video input (sequence of images), whereas \methodname{} learns to predict impact sounds only from a single image.

%% file: 03_method.tex
\section{\methodname{}: Curious Audiovisual Exploring Robot}

\methodname{} explores the visual and acoustic properties of objects in its surroundings in an efficient, self-supervised manner so as to build a rich audiovisual representation that enables downstream tasks (see Fig.~\ref{fig:method}).
Equipped with a novel, custom-built end-effector tool that enables the robot to hit different interaction points gently, \methodname{} takes a curiosity-driven approach that encourages exploration in the areas that promise to generate the most novel audiovisual signals. 
Our assumption is that visually distinct object regions are likely to generate different sounds, and \methodname{} leverages this insight to identify object parts with high visual uncertainty relative to the current dataset, providing fast coverage of diverse audiovisual signals.

At its core, \methodname{} relies on a KNN mapping visual embeddings to audio samples. The dependence of audio on environmental factors indicates that in domain data is likely important for accurate audio prediction. As such, rather than trying to train a generative model on a large dataset, \methodname{} instead aims to learn ground truth per object sounds in the relevant environment. In this way \methodname{}'s retrieval-based approach leverages the simplicity of KNNs as a tool for maximizing coverage of the audio domain in the context of its environment. 

To maximize the information gained with a new audiovisual sample, avoid repeatedly interacting with previously explored objects, and infer the most promising interaction points for downstream tasks, \methodname{} analyzes candidate samples on both the global and local level. \methodname{} evaluates the differences between the visual appearance of the candidate interaction points and previously interacted ones and selects the most uncertain one (the most distinct one), in a curiosity-inspired mechanism.
Conversely, given an audio sample, \methodname{} can also predict the most promising interaction point to recreate that sound by inverting the similarity-based search process. 
Both capabilities (uncertainty-based curious exploration and similarity-based exploitation) are necessary to 1) efficiently build a proper audiovisual representation and 2) achieve downstream tasks such as audiovisual material classification or reproducing a tune or an audio-demonstrated manipulation. 
In the following, we describe each of the components of \methodname{} in detail.

\subsection{Impact Tool}
\begin{figure}[t!]
\centering
\includegraphics[width=0.24\textwidth, trim={2cm 5cm 2cm 4cm},clip,valign=t]{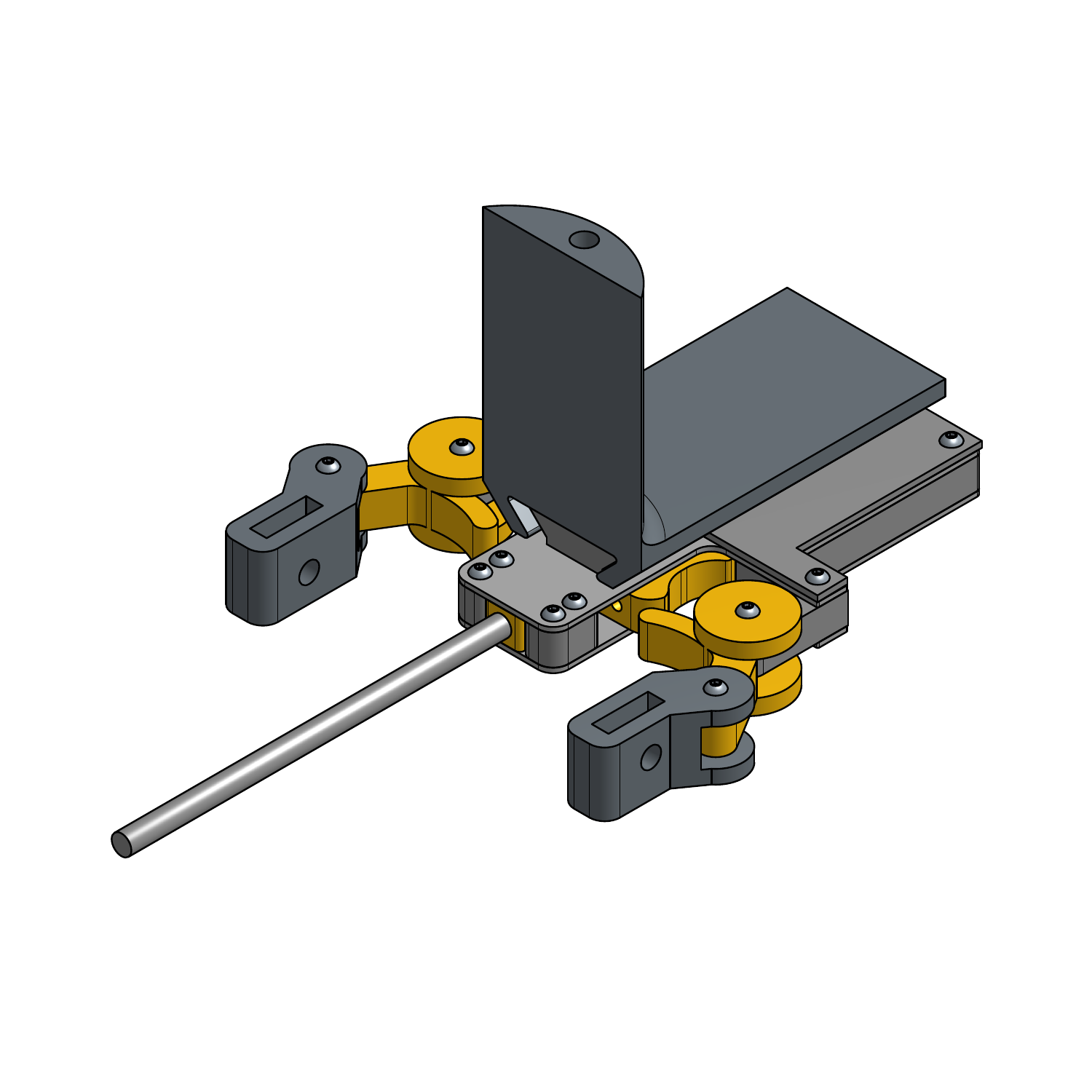}%
\includegraphics[width=0.24\textwidth,trim={2cm 0cm 0cm 0cm},clip,valign=t]{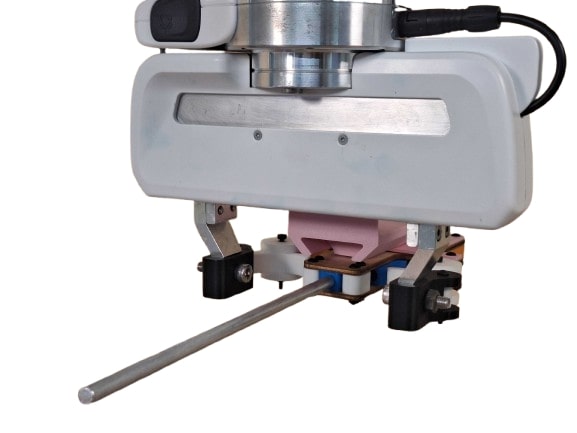}%
    \caption{\textbf{Robotic Impact Tool}. To generate consistent impact sounds, we designed a 3D-printed, spring-loaded impact tool that attaches to the robot's gripper. When the gripper closes, a cam-follower retracts the rod, building tension in the spring until the cam slips and the stored energy in the spring drives the metal rod forward, impacting the object. A directional microphone is mounted near the scene camera to record the resulting sound at 44.1 kHz. } 
    \label{fig:tool}
    \vspace{-1em}
\end{figure}

To generate consistent, informative impact sounds across diverse objects, we equip the robot with a mechanically actuated, single degree of freedom, 3D-printed, spring-loaded impact tool that can be grasped/mounted on a standard parallel gripper (Fig.~\ref{fig:tool}). 
Closing the gripper actuates a cam–follower that preloads the spring and releases a short, repeatable strike with a metal rod; a directional microphone mounted near the scene camera records the impact. The force of the impact is dictated by the thickness of the 3D-printed spring, which is set at 2mm after qualitative adjustments for consistent sample collection.  
This arrangement both standardizes excitation energy across trials and allows safe, gentle contacts at surface normals computed from the depth point cloud, so interaction is precise yet does not cause motion. Our data collection indicated that even after 1800 samples, the performance of this hitting implement did not measurably deteriorate.  
The result is audio that is comparable across objects and scenes, enabling a simple KNN model to perform surprisingly well in the low-data regime targeted by \methodname{}. 
Thanks to our custom tool, \methodname{} turns each interaction into a clean audiovisual training sample, powering both exploration and the downstream results in Sec.~\ref{sec:exp4}.

\subsection{\methodname{} Audiovisual Representation}

At its core, \methodname{} autonomously builds and maintains an increasing audiovisual representation of the interacted points that guides exploration and allows exploitation of the information for downstream tasks.
\methodname{}'s audiovisual representation is a growing set $\mathcal{S}$ of tuples $\{(x, \mathbf{f}(x), \mathbf{a}(x))\}$, where $x$ is an interaction point on an object, $\mathbf{f}(x)$ is a multi-scale \emph{visual} embedding of that point and its context, and \mbox{$\mathbf{a}(x)$} are \emph{audio} descriptors from the recorded impact. 
The representation is intentionally retrieval-friendly so it can be queried during exploration and downstream tasks.

Given a segmented scene, \methodname{} extracts visual features that preserve semantics and geometry at complementary scales. 
We use GroundedSAM (GSAM)~\cite{ren2024groundedsamassemblingopenworld, liu2024grounding, kirillov2023segment} to obtain mask embeddings for the \emph{object} and \emph{object part}, and a pre-trained ResNet to compute appearance features for the \emph{object crop}, \emph{part crop}, and a small \emph{local patch} centered at $x$.
We concatenate these into a single visual feature of the form:
\begin{equation}
\mathbf{f}(x)=\big[
    \mathbf{f}^{\text{SAM}}_{\text{obj}},\,
    \mathbf{f}^{\text{SAM}}_{\text{part}},\,
    \mathbf{f}^{\text{RN}}_{\text{obj}},\,
    \mathbf{f}^{\text{RN}}_{\text{part}},\,
    \mathbf{f}^{\text{RN}}_{\text{patch}}
\big]
\end{equation}

After a point is interacted with, \methodname{} acquires and creates an audio feature from the resulting impact waveform. 
First, \methodname{} level-normalizes and clips the audio signal around the strike; then, it computes mel-frequency cepstral coefficients (MFCCs) and stores them as audio features. 
MFCC features represent the shape of the spectral envelope of a sound, describing timbre instead of the intensity of the impact waveform. This makes them effective for identifying contact points independently of the strength of the interaction.
To evaluate and match audio to vision for downstream tasks (see Sec.~\ref{sec:exp4}), \methodname{} uses mel-cepstral distortion (MCD)~\cite{kubichek1993mel} between MFCC features as an audio-space distance metric between samples.  

\paragraph*{Aligning Audio and Visual Distances}
Distances in the concatenated visual space may not reflect acoustic similarity. 
We therefore learn per-component weights so that the weighted \emph{visual} distance correlates strongly with the \emph{audio} distance. 
Given two candidate points $x_i,x_j$, we define
\begin{equation}
\label{eq:visdist}
d_v(x_i,x_j)=\sum_{k} w_k\,\big\|\mathbf{f}_k(x_i)-\mathbf{f}_k(x_j)\big\|_2,
\end{equation}
where $k\in\{\text{SAM-obj},\text{SAM-part},\text{RN-obj},\text{RN-part},\text{RN-patch}\}$ indexes feature types. 
We fit the scalar weights $\{w_k\}$ by linear regression so that $d_v$ best predicts the corresponding MCD between paired impact audios.
To compute these weights before online exploration, we used synthetic data from the ObjectFolder~\cite{gao2023objectfolder} dataset. 
This alignment lets the system i) better retrieve plausible sounds from vision and most promising interaction points to recreate given audio by nearest neighbors, and ii) more accurately measure novelty for exploration directly in the visual space.

\subsection{Uncertainty-guided Exploration}
\methodname{} uses uncertainty to decide where to interact next so that each strike adds the most information. 
At each iteration, first generate per-object and per-part masks by prompting GroundedSAM \cite{ren2024groundedsamassemblingopenworld} with a ground-truth list of objects in the scene. The object masks are then projected onto the scene point cloud, and their novelty is computed with respect to the current audiovisual representation $\mathcal{S}$:
\begin{equation}
U(x)=\min_{s\in\mathcal{S}} d_v\big(x,s\big).
\end{equation}
The policy selects the farthest candidate $x^\star=\arg\max_x U(x)$, subject to a simple cycling rule over detected objects so that each instance is sampled before revisiting. 
Intuitively, this expands the frontier of audiovisual knowledge in the most visually unfamiliar regions first, while still allocating budget fairly across objects. 
In practice, combining cycling with curiosity delivers faster MCD drop than baselines (see Sec.~\ref{sec:exp4}-A, and Fig.~\ref{fig:exp1}), and it seeds a more discriminative embedding that boosts downstream classification (see Sec.~\ref{sec:exp4}-B, and Fig.~\ref{fig:exp2}).


\subsection{Exploiting the Representation for Downstream Tasks}

\methodname{} efficiently learns audiovisual correlations and gathers audiovisual data of its environment in a self-supervised, autonomous manner. 
Given this data, multiple downstream tasks are possible, leveraging the same nearest-neighbor-based retrieval machinery. 
Here, we study three in particular: audio generation, material classification, and audio-based imitation learning. 

\noindent\textbf{(i) Efficient exploration towards best audio prediction from vision.} Given a prospective hit, \methodname{} uses a nearest-neighbor retrieval under $d_v$ to obtain the audio exemplar that most closely matches the expected acoustics for that location. 
While this model yields modest audio generation capabilities compared to generative AI models, it proves effective and accurate in our low-data regime.
This retrieval methodology reveals the benefits of \methodname{}'s exploration strategy, which decreases MCD the fastest for a test set of held interaction points (Fig.~\ref{fig:exp1}).

\noindent\textbf{(ii) Material classification.} To evaluate this task, we train lightweight material prediction heads on visual, audio, or concatenated audiovisual embeddings, trained on a subset of interaction points data labeled manually with ground truth material labels. 
We observe that while \methodname{}'s visual features collected withe the curiosity-driven mechanism already provide strong predictive capabilities (as in humans) that are enhanced when audio is integrated in the audiovisual variant, yielding strong and sample-efficient accuracy (up to 87\% in our scenes; Fig.~\ref{fig:exp2}).

\noindent\textbf{(iii) Audio-guided imitation.} 
\methodname{} enables a robot to reproduce audio demonstrations by inverting the nearest neighbor computation: given a candidate audio sample, we can calculate the most similar audio in the dataset by finding the sample in the dataset with the lowest MCD to the query. 
Given this audio sample, we can retrieve the corresponding visually closest sample and identify the object part that most likely produced the sound. 
In our experiments, we perform two such imitation tasks: reproducing musical notes, and identifying objects involved in a pick-and-place manipulation performed by a human, closing the loop from \emph{listen} to \emph{do} (see Sec.~\ref{sec:exp3} and Sec.~\ref{sec:exp4}). 


%% file: 04_experiments.tex
\section{Experimental Evaluation}

\begin{figure*}[t!]

\centering
\begin{subfigure}[t]{0.99\textwidth}
\includegraphics[width=0.33\textwidth]{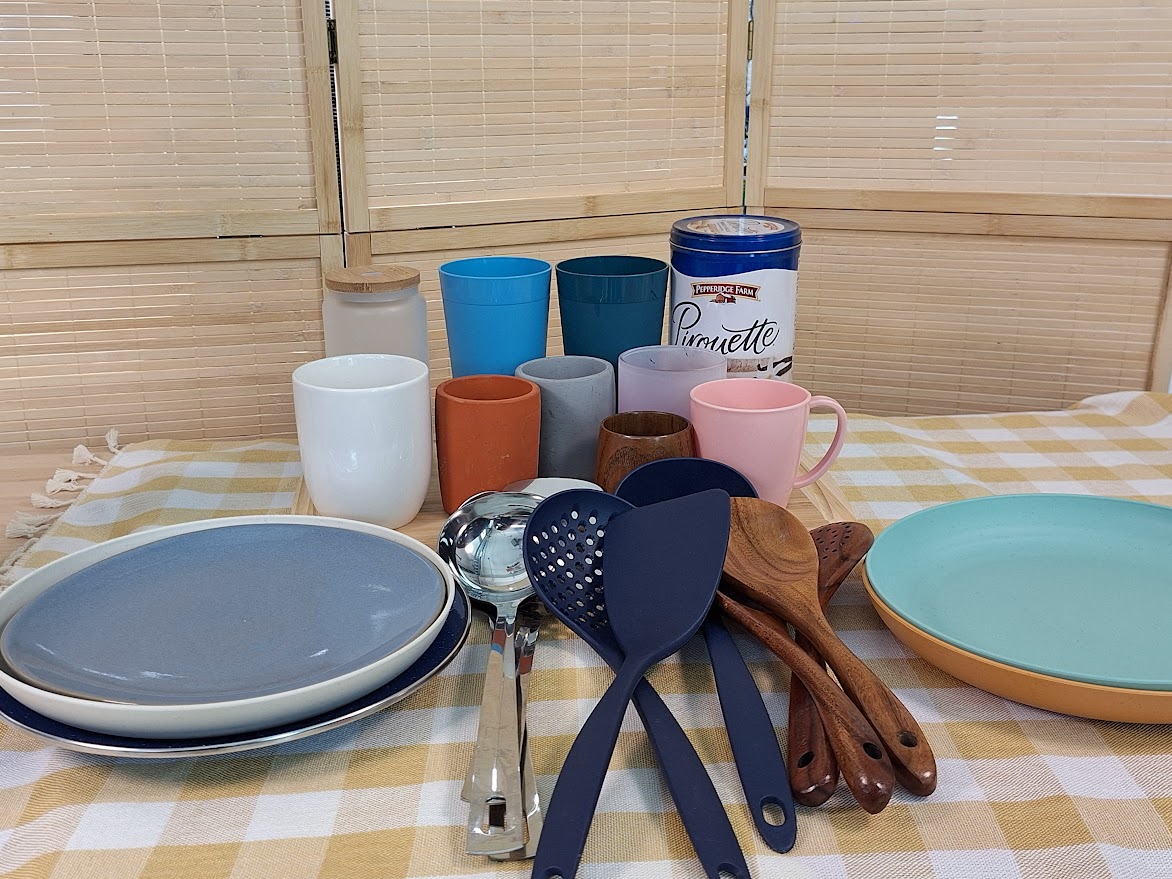}%
\hfill
\includegraphics[width=0.33\textwidth]{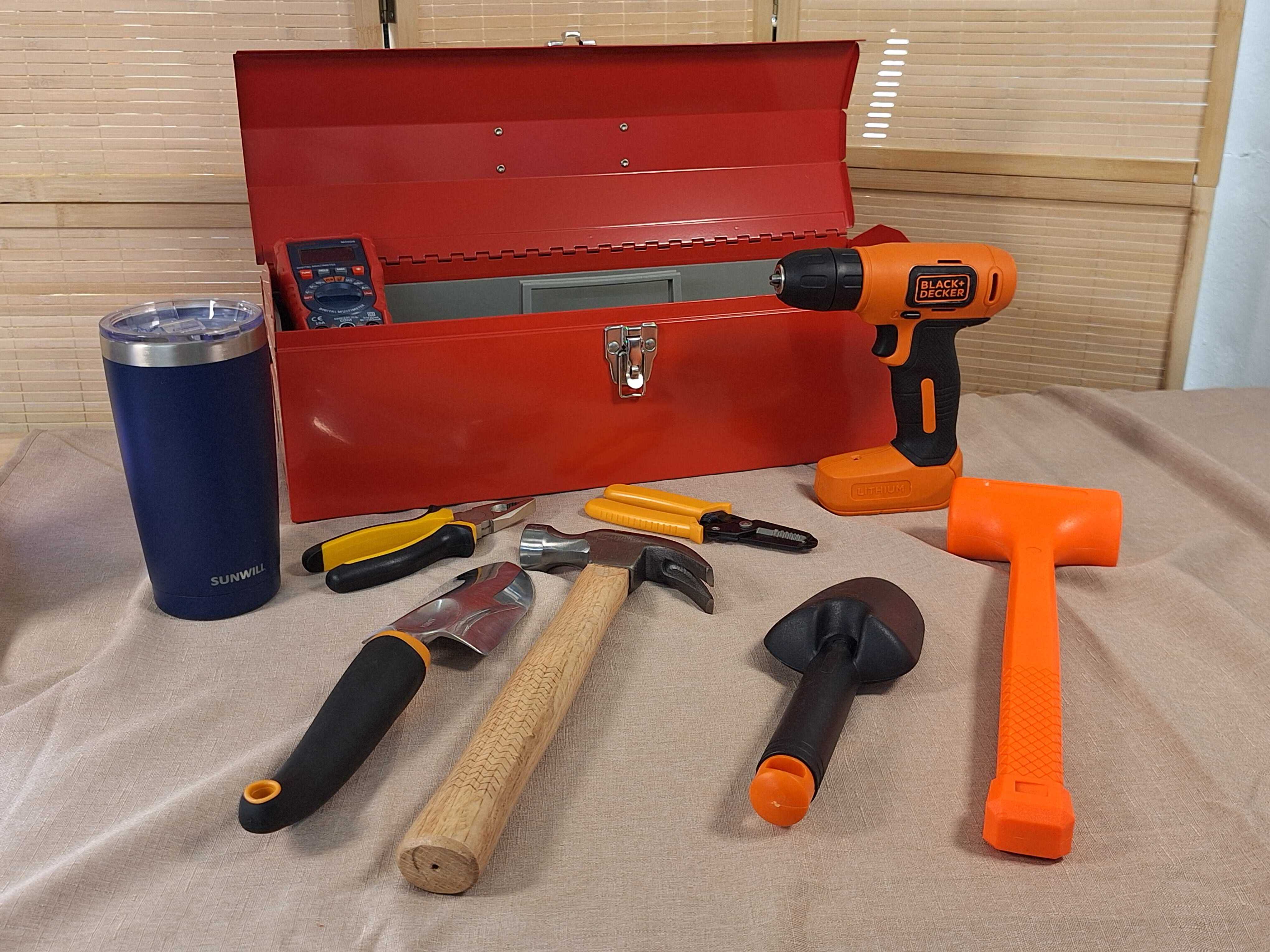}%
\hfill
\includegraphics[width=0.33\textwidth]{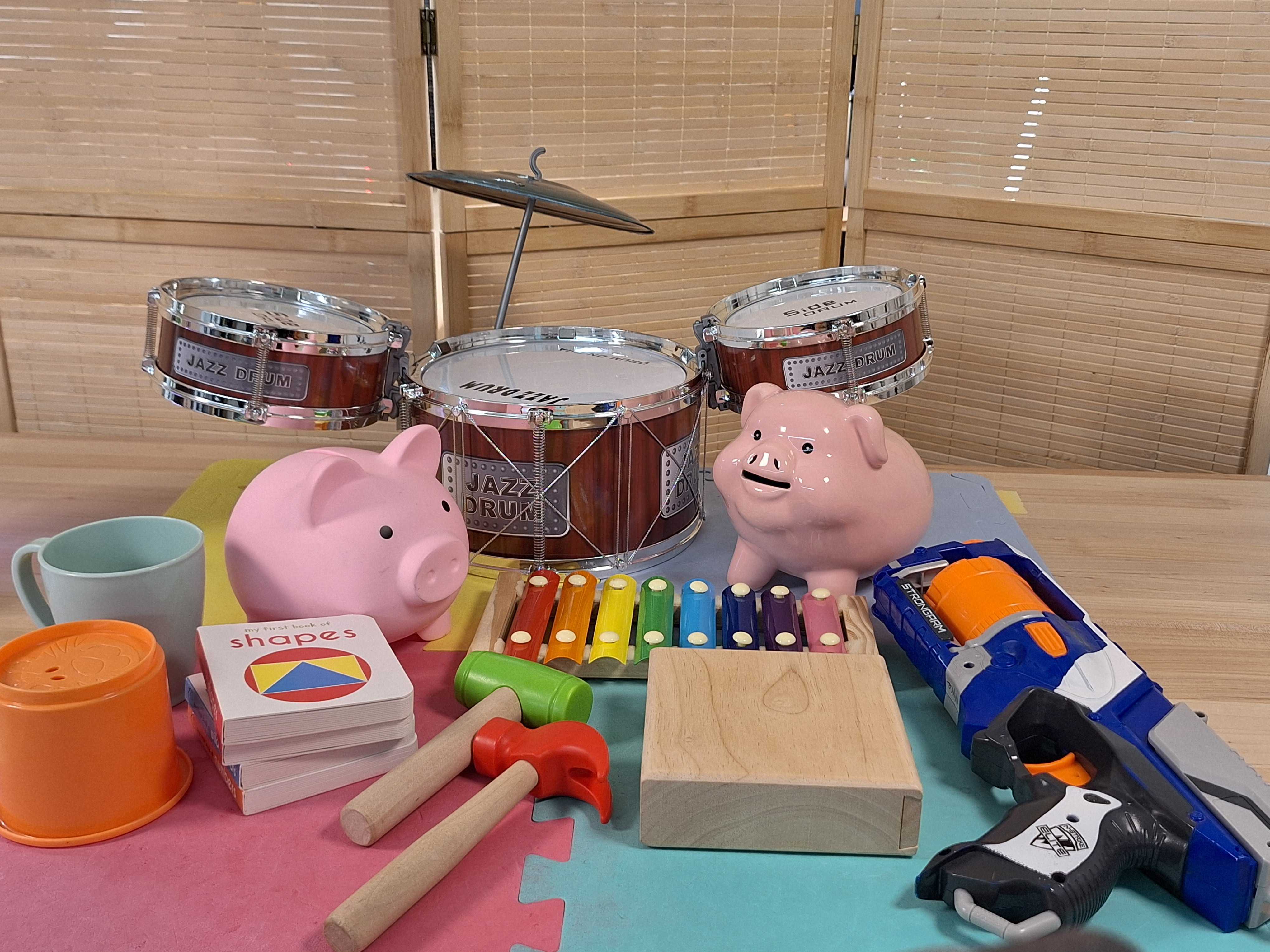}%
\caption{Our experiments take place across 3 environments: \texttt{kitchen}, \texttt{garage}, and \texttt{playroom}. }
\label{fig:envs}
\end{subfigure}
\begin{subfigure}[t]{0.99\textwidth}
\hspace*{-.4cm}
\includegraphics[width=\textwidth]{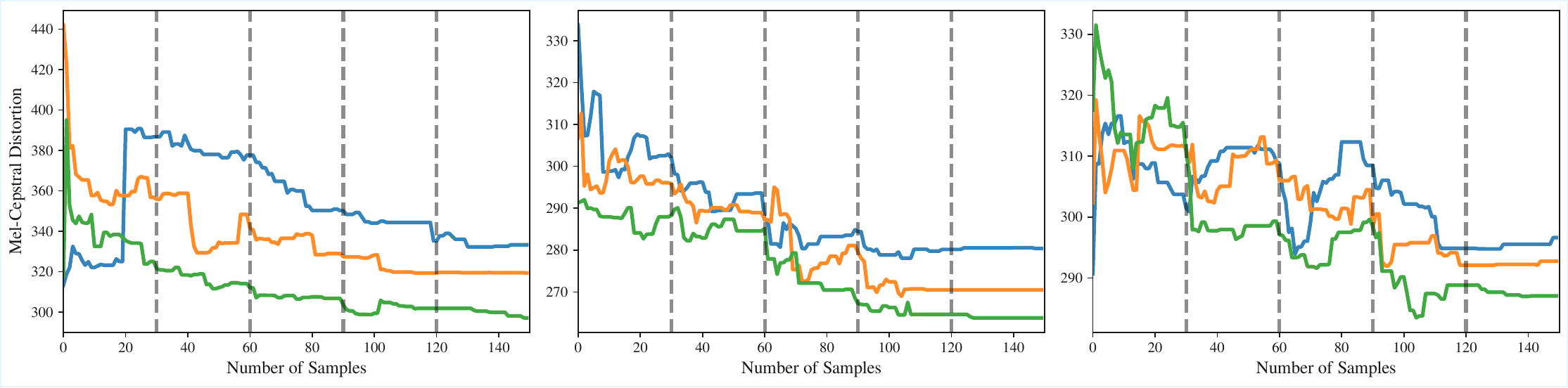}

\centering
\includegraphics[width=5cm]
{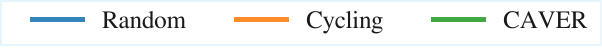}%
\caption{Audio prediction error as a function of samples collected for different audiovisual exploration approaches for the \texttt{kitchen}, \texttt{garage}, and \texttt{playroom}.}
\label{fig:exp1_results}
\end{subfigure}

\caption{\textbf{Curious Exploration for Audio Prediction}. The top row (a) visualizes the objects that comprise each environment. The bottom row (b) shows the average audio prediction error on a held-out test set, measured as the mel-cepstral distortion, as a function of the number of interaction points sampled by the robot. Vertical lines indicate scenes within an environment in which a subset of the objects appear. Each environment is plotted individually (left to right: \texttt{kitchen}, \texttt{garage}, and \texttt{playroom}). \methodname{} consistently achieves higher prediction accuracy more quickly than the naive exploration baselines via its curious exploration and audiovisual representation.}
\vspace{-1em}
    \label{fig:exp1}
    
\label{fig:scenes}
\end{figure*}


We evaluate \methodname{} exploration-exploitation capabilities on several downstream tasks that require learning and exploiting correlations between audio and visual appearance. 
We study four tasks in particular: predicting audio given visual input and hitting location, predicting material based on audio and vision, inferring musical notes played by a human, and identifying objects in a pick-and-place manipulation demonstrated by a human. 
Our evaluation is designed to test \methodname{}'s capabilities to efficiently explore the environment to build a rich and useful audiovisual representation and the accuracy of exploiting it for the four downstream tasks.

Using a fixed RGB-D camera, \methodname{} converts pointclouds into surface normal maps to help an exploration policy select and orient target impact poses for a Franka Research 3 robot. The system then executes collision-free paths via RRT \cite{lavalle2001rapidly} and Octomaps \cite{hornung2013octomap}, utilizing audio magnitude thresholds to automatically filter failed impacts before autonomously proceeding to the next sample.
To process audio, each impact recording is aligned by peak magnitude and clipped to a fixed window of 10 ms before and 100 ms after the peak. We apply per-clip amplitude normalization and compute features for MCD by resampling to 22.05 kHz and extracting the spectral envelopes (frame period 5.0 ms, FFT size 512). We convert these envelopes to 13th-order mel-cepstral coefficients (MCEP; $\alpha$=0.65) and flatten the resulting coefficient sequence into a fixed-length vector per interaction.
We measure end-to-end sampling time from timestamps between consecutively collected samples. This includes time spent on failed attempts and any resulting replanning/resampling. The average time per sample is 52.9 s across all environments ($\approx$ 68 samples/hour): $53.7 \pm 3.7$ s (Garage), $56.2 \pm 8.5$ s (Kitchen), and $48.6 \pm 3.1$ s (Playroom).

\begin{figure*}[t!]
    \centering
\includegraphics[width=0.99\linewidth]{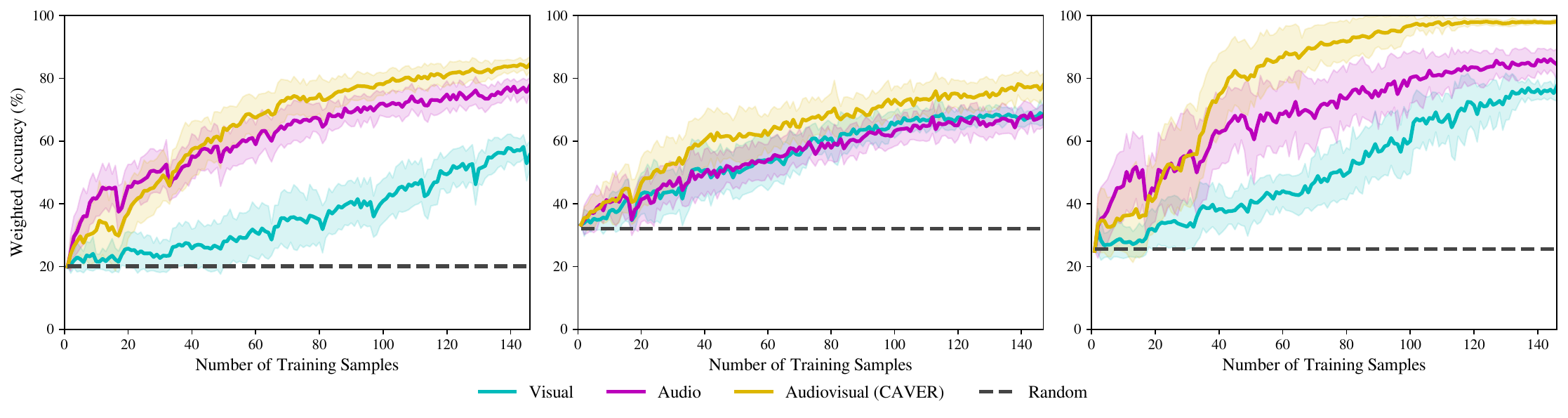}
    \caption{\textbf{Material Classification Results}. We measure the efficacy of \methodname{}'s unified audiovisual embeddings for material classification compared to visual embeddings alone, audio embeddings alone, and a random baseline. The plots show the class-balanced material classification accuracy on a held-out test set as a function of the number of interaction points sampled by the robot, aggregated over 20 runs. Performance is evaluated in the  \texttt{kitchen}, \texttt{garage}, and \texttt{playroom} environments, respectively. Incorporating both audio and visual inputs as in \methodname{}'s unified embedding is necessary to achieve strong material classification performance given ambiguous stimuli.}
    \label{fig:exp2}
    \vspace{-1em}
\end{figure*}

\noindent Our experiments aim to answer the following four questions: 

\subsection{Can \methodname{} enable a robot to efficiently gather data for training an audio prediction model?} 
\label{sec:exp1}
To assess \methodname{}'s exploration and audio generation capabilities, we designed an experiment in which the agent interacts with various household environments, sampling scenes to collect data autonomously. 
We designed three environments (\texttt{kitchen, garage}, and \texttt{playroom}), each consisting of five scenes with representative objects arranged on a table. 
Objects from each of the environments vary in size, color, texture, and materials, and are depicted in Fig.~\ref{fig:scenes}. 
The scenes are designed to test various aspects of the robotic system, including its ability to handle cluttered scenes, objects of similar visual appearance but different materials, and recurring objects across scenes. 
The scenes are experienced sequentially in order to test the model's ability to prioritize novel objects.
To evaluate performance, we manually curate a test set of impact sounds and corresponding hitting locations by sampling each part of each object that occurs in the scenes. 
To measure audio generation accuracy, we compute the mel-cepstral distortion (MCD) between the retrieved audio sample and the ground truth audio sample.  

In order to validate the effectiveness of our curiosity metric, we ablate \methodname{}'s point selection against the following baselines: 
\begin{itemize}
    \item \texttt{Random}: Randomly sample valid hitting locations by choosing points in the point cloud uniformly at random. 
    \item \texttt{Cycling}: Sample hitting locations by cycling through objects and choosing a random point on each object. This ablation of \methodname{} proceeds in an arbitrary order rather than prioritizing unfamiliar objects/points. 
\end{itemize}

Fig.~\ref{fig:exp1} shows the results for each household environment. Dashed vertical lines indicate the boundaries between scenes when the objects before the robot change. The test samples for each scene are added to the test set at the start of that scene. 
Audio prediction error naturally declines over time as the robot encounters more objects that are present in the withheld test set. 
We see that \methodname{} more quickly reduces the audio prediction error than the baselines, indicating that the curiosity-driven exploration indeed leads to more efficient learning of the audiovisual properties of the objects in each scene. This is especially apparent in later scenes when many of the objects have already appeared in earlier scenes and are thus already familiar to \methodname{}, whereas the baselines spend additional time sampling objects for which the uncertainty is already low. 

We also observe that \methodname{} struggles to retrieve quality audio samples when dealing with visually identical objects that have significantly different sound profiles (namely plastic white mug versus a ceramic white mug). This occurs when the same region in visual space maps to two separate regions of audio space, creating ambiguity as to what the object should sound like. This problem is inherent to audio-visual approaches, and instance-level object identification would be required to definitively solve it. In \ref{sec:exp4} we explore a workaround where objects can be sampled during tasks to make instance-level predictions.  

\begin{figure*}[t]
    \centering
    \begin{subfigure}[t]{0.49\textwidth}
    \includegraphics[width=1\textwidth]{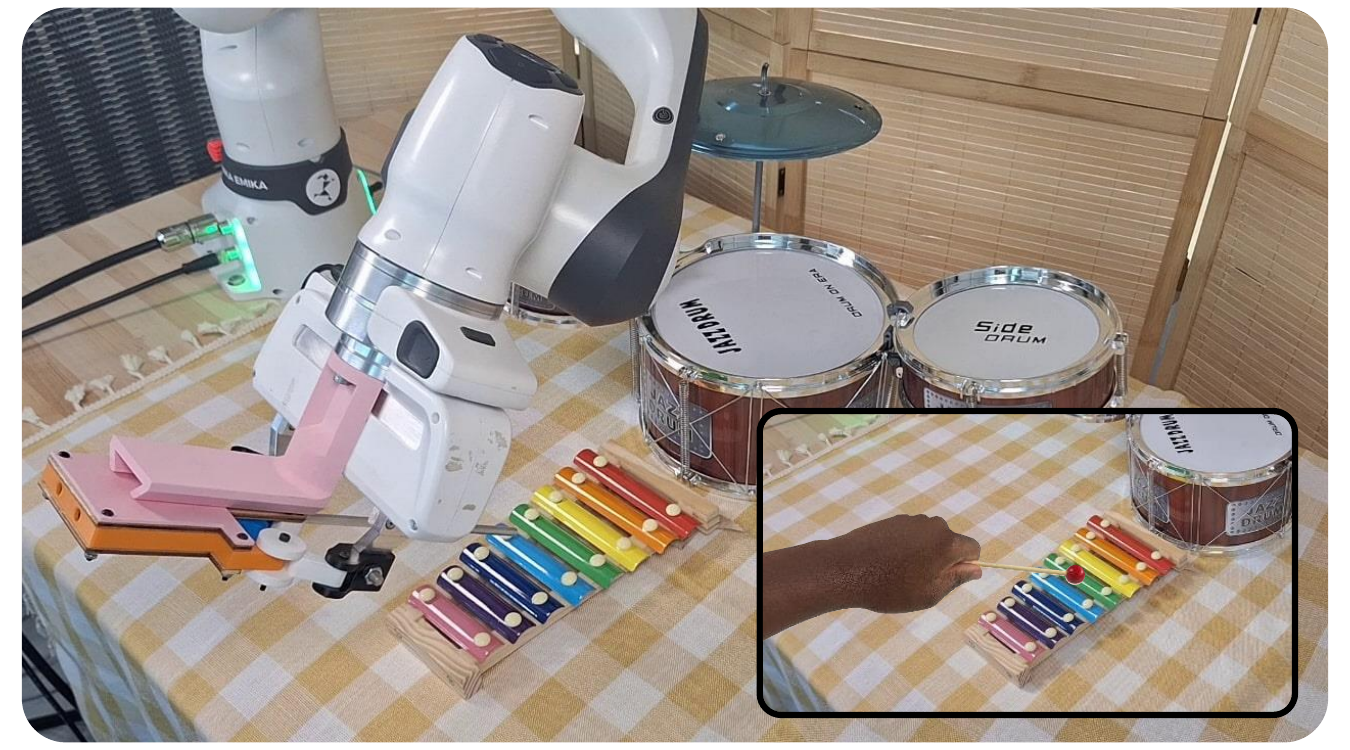}%
    \hfill
    \caption{CAVER replicating a song on the drums and xylophone.}
\label{fig:exp3_right}
    \end{subfigure}
        \begin{subfigure}[t]{0.49\textwidth}
    \includegraphics[width=1\textwidth]{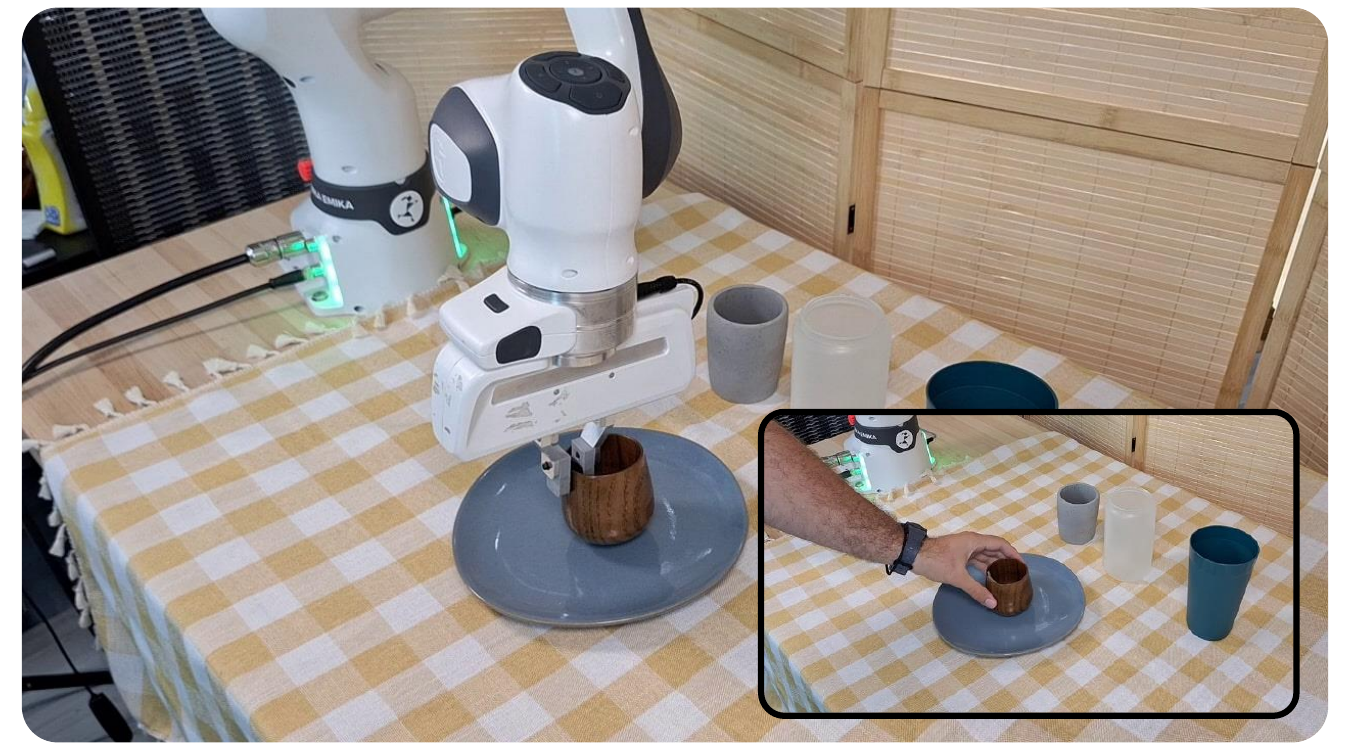}%
    \hfill
        \caption{CAVER imitating previously unheard object interactions.}
\label{fig:exp3_left}
    \end{subfigure}

    \hfill
    \caption{\textbf{Audio Imitation with \methodname{}}. \methodname{}'s audiovisual representation enables imitation of solely auditory demonstrations. In (a), \methodname{} plays a song by ear after exploring the instruments autonomously. In (b), the robot replicates unseen manipulation actions by predicting what objects sound like together (ie, combining the audio from a wood cup sample and a ceramic plate sample) and comparing them to the observed sound. The ability to reason about multi-object interaction sounds gives \methodname{} the flexibility to be applied to diverse downstream tasks.}
    \label{fig:exp3}
    \vspace{-1em}
\end{figure*}

\subsection{Can \methodname{} enable efficient learning of object material prediction given audio data?}
\label{sec:exp2}

In a second experiment, we evaluate \methodname{}'s embeddings in a material classification task. The setup is identical to Experiment 1 above; however, we evaluate material classification performance rather than audio generation error. 
Ground truth material labels for each sample are provided by a human annotator. We compare three variants of  MLP-based material classification models: \texttt{vision-only} uses only \methodname{}'s visual embeddings; \texttt{audio-only} uses audio —featurized using MFCC, Chroma, and Mel Spectrogram\cite{aristorenas2024machinelearningframeworkaudiobased}— to predict the material; and \texttt{audiovisual} uses the concatenated latent features of both modalities during its forward pass. The visual features are encoded using MLPs and audio features with a CNN. The complete list of material classes consists of ceramic, glass, metal, plastic, rubber, and wood. 

To evaluate the effectiveness of \methodname{}'s visual embeddings and the importance of audiovisual data in material classification, we first evaluate material classification accuracy on data from \methodname{}'s exploration described above in Sec.~\ref{sec:exp1}. The order in which the data is provided to the model is randomized across 20 runs for robustness. This procedure is repeated across each environment. Fig.~\ref{fig:exp2} showcases the classification performance results for each model as a function of samples collected. We observe that the \texttt{audiovisual} model has the strongest performance across all three environments, demonstrating the utility of both combined audio and visual features when classifying materials. In addition, \texttt{audio-only} outperforms \texttt{vision-only} in environments where there are many objects of the same type, demonstrating the utility of the curiously collected data.

\subsection{Can \methodname{}'s audiovisual representation enable an agent to reproduce musical sounds? }
\label{sec:exp3}
Human expert musicians are capable of reproducing a tune on their instrument upon hearing it once. In this experiment, we evaluate \methodname{}'s ability on this same imitation task. 
We designed two tasks to evaluate this capability: playing the drums, and playing a xylophone. The instruments are pictured in Fig.~\ref{fig:exp3}. For each instrument, a test set of sounds is generated by a human demonstrator. We evaluate \methodname{}'s ability to infer the correct musical notes and compare its performance to a human baseline. The human is allowed to play with each instrument for a few minutes before being played test sounds and asked to guess the correct note. \methodname{} first explores the scene to build its KNN model. To reproduce a given sound, we sample points in the environment and predict their corresponding audio. The predicted audio with the lowest MCD distance to the test sound is selected as \methodname{}'s prediction. 
\methodname{} is able to infer the correct xylophone key {64\%} of the time, and the correct drum hit with {68\%} accuracy. These results demonstrate the effectiveness of \methodname{}'s visual representation and sample retrieval. Please see the supplement for a demonstration of \methodname{}'s musical ability. 




\subsection{Can \methodname{} enable activity recognition by inferring objects that are interacted with based on sound alone? }
\label{sec:exp4}
A final capability enabled by \methodname{} is imitation of a manipulation behavior based on sound alone: given a test sound, \methodname{} enables the robot to infer which object(s) were involved in the production of this sound. We evaluate this capability on a pick-and-place task. A human picks an object of varying material and places it on a plate of known material. The robot's task is to infer which object was placed and reproduce the demonstrated behavior. \methodname{} interacts with the picked objects and placement objects for a period of time before the evaluation begins. We assume that the placement object is known and only the picked object needs to be determined. To estimate which objects were involved in the manipulation, \methodname{} synthesizes a combined sound by superimposing the predicted sounds of the picked and placement objects. As above, the interaction point yielding the lowest MCD to the test audio is selected and that object is picked.
We find that \methodname{} is able to correctly infer the manipulated object 42\% of the time, whereas the average human performance is only 27\%. 

%% file: 05_conclusion.tex
\section{Conclusion and Limitations}
We introduced \methodname{}, a robot capable of efficiently and autonomously learning the audiovisual properties of objects. By combining a nearest-neighbor retrieval system with uncertainty-driven exploration, \methodname{} can generate impact sounds conditioned on visual input and accelerate the acquisition of acoustic properties compared to naive sampling strategies. This enables robots to rapidly acquire auditory knowledge that supports downstream tasks such as sound prediction, material classification, and activity recognition, as demonstrated in our comprehensive evaluation.
While \methodname{} performs well in all our experiments, it presents several limitations that we plan to explore: 1) \methodname{} does not yet capture continuous sounds (e.g., dragging), 2) it completely separates exploration from inference, 3) it lacks mechanisms to filter background noise that could distort the learned audiovisual representation 4) failure states (unsampleable points, mechanical failure during sampling, etc.) are discarded and 5) \methodname{} assumes a ground truth list of objects for visual segmentation. Addressing these challenges will enable larger-scale deployments in more open-ended environments, such as homes with diverse objects and ambient conditions, where the benefits of uncertainty-guided exploration are especially critical. We also hope further exploration of the audio-visual field will yield appropriate baselines against which to benchmark \methodname{}'s performance.
Overall, we view \methodname{} as a step toward equipping robots with richer multimodal understanding, bridging vision and audition to support new capabilities and more adaptive, perceptive, and interactive behavior in the real world.

\section{Acknowledgements}
We would like to thank Dr. David Harwath for his guidance in audio representation and analysis techniques. This project has been partially funded by Good Systems, an initiative by the University of Texas at Austin.